\documentclass[sigconf]{acmart}
\AtBeginDocument{%
  }

\copyrightyear{2026}
\acmYear{2026}
\setcopyright{cc}
\setcctype{by}
\acmConference[IAIT 2026]{14th International Conference on Advances in Information Technology}{June 17--19, 2026}{Bangkok, Thailand}
\acmBooktitle{14th International Conference on Advances in Information Technology (IAIT 2026), June 17--19, 2026, Bangkok, Thailand}
\acmDOI{10.1145/3816713.3818806}
\acmISBN{979-8-4007-2436-7/2026/06}

\acmSubmissionID{6030}

\usepackage{multirow}
\usepackage{array}
\usepackage{tabularx}
\usepackage{subcaption}

\begin{document}

\title{Prompt Compression in Diffusion Large Language Models: Evaluating LLMLingua-2 on LLaDA}

\author{Sterling Huang}
\authornote{Equal contribution.}
\email{sterling.huang@mail.utoronto.ca}
\affiliation{%
  \institution{University of Toronto}
  \city{Toronto}
  \country{Canada}
}

\author{Abigayle Brown}
\authornotemark[1]
\email{abbey.brown@mail.utoronto.ca}
\affiliation{%
  \institution{University of Toronto}
  \city{Toronto}
  \country{Canada}
}

\author{Jiyoo Noh}
\authornotemark[1]
\email{jiyoo.noh@mail.utoronto.ca}
\affiliation{%
  \institution{University of Toronto}
  \city{Toronto}
  \country{Canada}
}

\author{Jiakang Xu}
\authornotemark[1]
\email{amyj.xu@mail.utoronto.ca}
\affiliation{%
  \institution{University of Toronto}
  \city{Toronto}
  \country{Canada}
}

\author{Wantong Huo}
\authornotemark[1]
\email{vanessa.huo@mail.utoronto.ca}
\affiliation{%
  \institution{University of Toronto}
  \city{Toronto}
  \country{Canada}
}

\author{Kaung Myat Kyaw}
\authornotemark[1]
\email{kaungmyat.kyaw@kmutt.ac.th}
\affiliation{%
  \institution{King Mongkut's University of Technology Thonburi}
  \city{Bangkok}
  \country{Thailand}
}

\author{Jonathan H. Chan}
\email{jonathan@sit.kmutt.ac.th}
\affiliation{%
  \institution{King Mongkut's University of Technology Thonburi}
  \city{Bangkok}
  \country{Thailand}
}


\renewcommand{\shortauthors}{Huang et al.}

\begin{abstract}
Prompt compression reduces inference cost and context length in large language models, but prior evaluations focus mainly on autoregressive architectures. This study examines whether LLMLingua-2 transfers effectively to diffusion large language models (DLLMs), specifically LLaDA-8B-Instruct. We evaluate GSM8K, DUC2004, and ShareGPT using 250 prompts per dataset at an approximate 50\% compression ratio, covering mathematical reasoning, prompt reconstruction, and summarization. Outputs from original, compressed, and reconstructed prompts are compared using exact-match accuracy, BLEU, ROUGE, and BERTScore. Results show that high semantic preservation does not necessarily ensure stable downstream behavior in diffusion models. Summarization remains relatively robust, while mathematical reasoning degrades substantially despite high semantic similarity. Reconstruction further shows that semantically similar prompts may omit reasoning-critical information needed for stable denoising. Overall, compression failures are mainly driven by information omission rather than semantic drift, suggesting that autoregressive prompt compression methods may not transfer uniformly to DLLMs. These findings motivate diffusion-aware compression strategies. The source code is available at: \url{https://github.com/axxx129-ops/DLLM-Compression}.

\end{abstract}


\maketitle

\section{Introduction}
Prompt compression has emerged as a practical technique for reducing input length, inference latency, and computational cost in large language models (LLMs). Existing methods, including LLMLingua \cite{jiang2023llmlingua}, LongLLMLingua \cite{jiang2023longllmlingua}, and LLMLingua-2 \cite{pan2024llmlingua2}, selectively retain or remove prompt tokens based on estimated task relevance, while attempting to preserve downstream generation quality under shortened contexts. Recent empirical studies further demonstrate that compression effectiveness varies across tasks, target models, and evaluation metrics~\cite{zhang2025empirical}. However, these prior works have focused mostly on autoregressive large language models (AR LLMs), leaving the behavior of prompt compression in alternative generative architectures largely unexplored.

This limitation is particularly relevant for diffusion large language models (DLLMs). Unlike autoregressive LLMs \cite{grattafiori2024llama,yang2025qwen3,liu2024deepseek,ghalandari2022efficient,team2025kimi,laban2021keep} which generate text through left-to-right next-token prediction, DLLMs use a denoising process over masked tokens. This architecture enables bidirectional attention and partially parallel decoding, addressing limitations of strictly sequential generation. LLaDA~\cite{nie2026large}, for example, performs generation through a forward masking process followed by iterative reverse denoising to predict masked tokens.
These architectural differences suggest that prompt compression behaves differently in DLLMs than in autoregressive models. In autoregressive generation, compressed prompts primarily affect the conditioning prefix used for next-token prediction. In contrast, DLLMs rely on prompt information throughout the denoising process. Consequently, tokens discarded under autoregressive compression heuristics may still encode entities, numerical values, discourse structure, or contextual dependencies that influence denoising trajectories and output stability in DLLMs. This distinction raises an important question: do prompt compression methods developed for autoregressive models preserve downstream behavior when applied to diffusion-based language models?

In this work, we evaluate the transferability of prompt compression to DLLMs using LLMLingua-2 \cite{pan2024llmlingua2} as the compression framework and LLaDA \cite{nie2026large} as the target model. LLMLingua-2 provides a natural choice for this investigation because it formulates compression as a token classification problem, leverages bidirectional contextual information, and is designed to operate in a task-agnostic manner. Similarly, LLaDA serves as an appropriate target architecture because it extends masked diffusion large language modeling to an 8B-parameter instruction-following LLM. 

Importantly, this study distinguishes between prompt-level semantic preservation and downstream task behavior. This distinction is critical because a compressed or reconstructed prompt may remain semantically similar to the original while still losing specific entities, quantities, discourse relations, or contextual dependencies required to produce consistent outputs. To evaluate this phenomenon, we compare multiple behavioral settings, including original prompts, compressed prompts, reconstructed prompts, and responses generated from reconstructed prompts when reintroduced into LLaDA. We evaluate whether diffusion large language models may respond differently to compressed conditioning signals than autoregressive architectures.
\section{Related Work}
\subsection{Prompt Compression}
Prompt compression aims to reduce prompt length while preserving downstream generation quality in large language models (LLMs). LLMLingua introduced a coarse-to-fine prompt compression framework that estimates token importance with a smaller language model and uses budget control to reduce prompt length while preserving essential content \cite{jiang2023llmlingua}. LongLLMLingua extends compression to long-context settings by addressing cost, position bias, and the density of key information in retrieval-style prompts~\cite{jiang2023longllmlingua}. LLMLingua-2 reframes prompt compression as token classification and uses data distillation from stronger models~\cite{pan2024llmlingua2}. The framework is built on XLM-RoBERTa-large ~\cite{conneau2020unsupervised}, enabling efficient bidirectional contextual encoding during token selection. This makes LLMLingua-2 less directly tied to autoregressive decoding behavior than entropy-based compressors, which motivates its use as the compressor in this study. Additionally, prior empirical evaluations demonstrate that LLMLingua-2 achieves strong overall compression performance on autoregressive LLMs~\cite{zhang2025empirical}. 

The closest empirical reference point is the large-scale prompt-compression study of Zhang et al.~\cite{zhang2025empirical}, which evaluates six compression methods across 13 datasets and multiple autoregressive LLMs. Their results show that compression can affect generation quality, hallucination behavior, and word-omission patterns. We use it as a methodological reference, but we do not claim a controlled autoregressive comparison because our experiments do not rerun matched autoregressive baselines under identical prompts, decoding settings, and evaluation pipelines.
\begin{figure*}[t]
    \centering
    \includegraphics[width=0.9\textwidth]{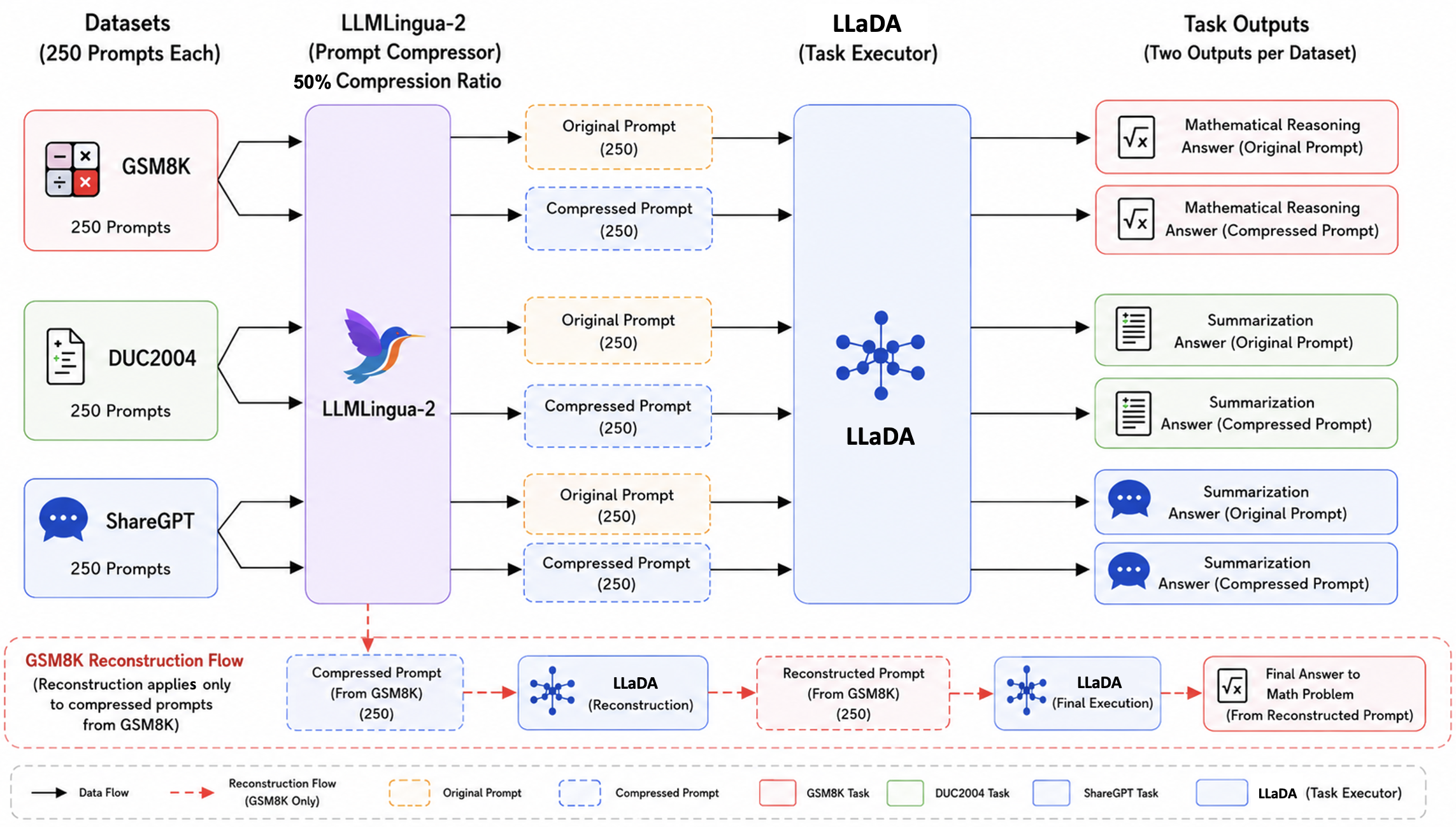}
    \Description{A pipeline diagram showing prompt collection, LLMLingua-2 compression, LLaDA generation or reconstruction, and evaluation against original, compressed, reconstructed, and downstream reasoning outputs.}
    \caption{Experimental pipeline. Prompts from GSM8K, DUC2004, and ShareGPT are compressed with LLMLingua-2 and evaluated with LLaDA under original-prompt, compressed-prompt, reconstructed-prompt, and downstream reasoning conditions.}
    \label{fig:pipeline}
\end{figure*}

\subsection{Diffusion Large Language Models}
DLLMs introduce a forward process that progressively injects noise into data within the discrete domain, while the training objective focuses on learning the distribution of the corresponding reverse process \cite{nie2026large}.
Within this framework, the forward noising process is characterized by the progressive substitution of tokens with a special \texttt{[MASK]} token. 
Let $\mathbf{x}_0 = (x_0^1, \dots, x_0^L)$ denote a text sequence of $L$ tokens sampled from the data distribution $p_{\text{data}}(\mathbf{x}_0)$, where each token is represented as a one-hot vector over a vocabulary of size $V$.
The forward process corrupts the original sequence $\mathbf{x_0}$ by randomly masking tokens, a mechanism governed by the following transition probability:
\begin{equation}
    q_{t \mid 0}(\mathbf{x}_t \mid \mathbf{x}_0) = \prod_{i=1}^L \text{Cat}\left(x_t^i; \alpha_t\delta_{x_0^i} + (1-\alpha_t)\delta_{\text{MASK}}\right).
\end{equation}
Here, $t \in [0,T]$ represents the diffusion time (or masking level), which regulates the interpolation between the original data $\mathbf{x}_0$ (at $t=0$) and the fully masked sequence (at $t=T$) via a predefined noise scheduler $\alpha_t$, such that $\alpha_0 \approx 1$ and $\alpha_T \approx 0$.
The term $\delta$ signifies a Dirac delta distribution, acting as a one-hot probability distribution. Specifically, $\delta_{x_0^i}$ and $\delta_{\text{MASK}}$ are one-hot distributions assigning probability 1 to the original token $x_0^i$ and the \texttt{[MASK]} token, respectively. 
The reverse process is designed to iteratively recover the values of masked tokens, initializing from a fully masked sequence $\mathbf{x}_T = [\texttt{MASK}, \dots, \texttt{MASK}]$.
For any time steps $0 \le s < t \le T$, the reverse process factorizes over the sequence:
\begin{equation}
    q_{s \mid t}(\mathbf{x}_s \mid  \mathbf{x}_t, \mathbf{x}_0) = \prod_{i=1}^L q_{s \mid t}(x_s^i \mid \mathbf{x}_t, \mathbf{x}_0).
\end{equation}
The generative model $p_\theta(\mathbf{x}_{s} \mid \mathbf{x}_t)$, parameterized by $\theta$, is trained to approximate this true posterior $q_{s \mid t}(\mathbf{x}_s \mid \mathbf{x}_t, \mathbf{x}_0)$. During inference, this learned reverse process is utilized for conditional generation. Given a prompt $\mathbf{c}$, generation initiates by appending \texttt{[MASK]} tokens to form the initial noisy state $\mathbf{x}_T = (\mathbf{c}, \texttt{[MASK]}, \dots, \texttt{[MASK]})$. Proceeding through reverse time steps $t=T\to0$, the model predicts the probability distributions for all masked positions simultaneously. At each step, a subset of tokens is unmasked, and this process iterates until the sequence is fully unmasked.

\subsection{Evaluation Metrics}
Lexical metrics such as BLEU \cite{papineni2002bleu} and ROUGE \cite{lin2004rouge} are useful diagnostics but can be brittle when outputs paraphrase the same content. BERTScore \cite{zhang2019bertscore} computes token similarity using contextual embeddings and has been proposed as a semantic text-generation metric~\cite{zhang2019bertscore}. In this paper, we use BERTScore as the only semantic preservation diagnostic and do not treat it as a replacement for task accuracy, exact matching, or qualitative failure analysis.

\section{Experimental Design}
\subsection{Method}
The objective of this study is to evaluate whether LLMLingua-2, a prompt compression framework designed for autoregressive large language models, transfers effectively to the diffusion large language model LLaDA. We examine the impact of compressed prompts on mathematical reasoning, reconstruction, and summarization quality. We evaluate three datasets with different task structures: GSM8K~\cite{cobbe2021gsm8k} for mathematical reasoning and prompt reconstruction, and DUC2004~\cite{duc2004} and ShareGPT~\cite{chen2024sharegpt4v} for summarization. For each dataset, 250 prompts are sampled. 

\subsection{Prompt Compression Pipeline}
The full experimental pipeline is shown in Figure \ref{fig:pipeline}. All experiments were conducted using LLaDA-8B-Instruct\footnote{\url{https://huggingface.co/GSAI-ML/LLaDA-8B-Instruct}} under identical decoding settings to isolate the effect of prompt compression. For each dataset, prompts were first compressed using LLMLingua-2 at approximately 50\% compression, which is the default compression ratio used in the empirical study~\cite{zhang2025empirical}, retaining about 50\% of the original prompt tokens. Both original and compressed prompts were evaluated using LLaDA under two conditions:
\begin{enumerate}
    \item original prompt, $p_i^{\mathrm{orig}}$;
    \item compressed prompt, $p_i^{\mathrm{comp}}$, generated using LLMLingua-2.
\end{enumerate}

For mathematical reasoning tasks, responses were generated from both prompt variants:
\begin{equation}
 r_i^{\mathrm{orig}} = \mathrm{LLaDA}(p_i^{\mathrm{orig}}), \quad
 r_i^{\mathrm{comp}} = \mathrm{LLaDA}(p_i^{\mathrm{comp}}).
\end{equation}

For reconstruction tasks, the original prompt was reconstructed from the compressed prompt:
\begin{equation}
 \hat{p}_i^{\mathrm{orig}} = \mathrm{LLaDA}_{\mathrm{recon}}(p_i^{\mathrm{comp}}).
\end{equation}
The reconstructed prompt was then re-evaluated for math reasoning, as shown in the red dashed-bordered region of Figure \ref{fig:pipeline}:
\begin{equation}
 \hat{r}_i = \mathrm{LLaDA}(\hat{p}_i^{\mathrm{orig}}).
\end{equation}
For summarization tasks, outputs were generated from both prompt conditions:
\begin{equation}
 s_i^{\mathrm{orig}} = \mathrm{LLaDA}(p_i^{\mathrm{orig}}), \quad
 s_i^{\mathrm{comp}} = \mathrm{LLaDA}(p_i^{\mathrm{comp}}).
\end{equation}
All experiments use fixed generation hyperparameters: 128 denoising steps, maximum generation length of 128 tokens, block length 32, temperature 0.0, and classifier-free guidance scale 0.0.

\begin{table*}[t]
\caption{Task setup and primary evaluation comparisons.}
\label{tab:task_setup}
\centering
\begin{tabularx}{\textwidth}{lllX}
\toprule
Dataset & Task & Prompt Conditions & Primary Comparison \\
\midrule
GSM8K & Mathematical reasoning & Original and compressed & Original-prompt answer vs. ground truth; compressed-prompt answer vs. ground truth; original-prompt answer vs. compressed-prompt answer \\
GSM8K & Prompt reconstruction & Compressed to reconstructed & Reconstructed prompt vs. original prompt \\
GSM8K & Reconstruction preservation & Reconstructed prompt & Reconstructed-prompt answer vs. original-prompt answer \\
DUC2004 & Formal summarization & Original and compressed & Compressed-prompt summary vs. original-prompt summary \\
ShareGPT & Conversational summarization & Original and compressed & Compressed-prompt summary vs. original-prompt summary \\
\bottomrule
\end{tabularx}
\end{table*}

\subsection{Tasks and Datasets}
In our study on prompt compression for diffusion large language models, we evaluated three tasks: mathematical reasoning, prompt reconstruction, and summarization. We use a subset of the datasets with 250 samples from the original study, selecting the GSM8K\footnote{\url{https://huggingface.co/datasets/openai/gsm8k}} ~\cite{cobbe2021gsm8k}, DUC2004\footnote{\url{https://duc.nist.gov/duc2004/}} ~\cite{duc2004}, and ShareGPT\footnote{\url{https://huggingface.co/datasets/liyucheng/sharegpt-500}} ~\cite{li2023compressing} datasets for reconstruction, mathematical reasoning, and summarization tasks. Although the original study~\cite{zhang2025empirical} evaluated 18 datasets, our experiments were limited to these three datasets due to computational constraints, as well as the lack of long-context and multimodal input support in the evaluated DLLMs. For instance, BBC \footnote{\url{https://huggingface.co/datasets/RealTimeData/bbc_news_march_2023}}\cite{li2023compressing} is too long to fit into LLaDA due to the lack of long context capability.

\noindent\textbf{GSM8K: Mathematical Reasoning.}
GSM8K contains grade-school-level mathematical word problems that require multi-step reasoning to derive final numerical answers. The task involves zero-shot prompting the model to solve the mathematical problem from both the original and compressed prompt and comparing similarity across three numerical answers: original-prompt answers $r_i^{\mathrm{orig}}$, compressed-prompt answers $r_i^{\mathrm{comp}}$, and human-labeled ground-truth answers $y_i$. The primary objective is to determine whether compressed prompts preserve reasoning fidelity reflected by answer correctness. 

\noindent\textbf{GSM8K: Prompt Reconstruction.}
Using the same GSM8K dataset, we study prompt reconstruction by prompting the model to reconstruct the original prompt from the compressed prompt generated by LLMLingua-2. The reconstructed prompts are then evaluated for semantic similarity against the original prompts using semantic preservation metrics. To further evaluate whether task-relevant information is preserved during reconstruction, the reconstructed prompts are fed back into the model and used again for mathematical reasoning.

\noindent\textbf{DUC2004 and ShareGPT: Summarization.}
We evaluate output preservation on two text-generation datasets. DUC2004 was selected as a formal news-style summarization dataset, while ShareGPT was selected to represent conversational summarization. The task involves generating summaries from both the original and compressed prompts and comparing the similarity between these two outputs to evaluate semantic preservation under compression. Table~\ref{tab:task_setup} summarizes the task setup and primary evaluation comparisons.

\subsection{Evaluation Metrics}
For GSM8K mathematical reasoning, we report exact-match agreement between numerical answers produced from the original and compressed prompts in Table \ref{tab:ar_vs_dllm}. We also report exact-match accuracy against ground-truth answers for both the original-prompt and compressed-prompt settings in Table \ref{tab:math_partitions}. ShareGPT summarization is not included in Table \ref{tab:ar_vs_dllm} because the autoregressive baseline study \cite{zhang2025empirical} did not report ShareGPT summarization results.

Reasoning trajectories are categorized into five groups based on equality relationships among the ground truth, the original-prompt response, and the compressed-prompt response: \textit{all same}, \textit{all different}, \textit{original-prompt response matches compressed-prompt response only}, \textit{original-prompt response matches ground truth only}, and \textit{compressed-prompt response matches ground truth only}. These categories capture patterns such as preserved behavior, compression-induced failure, compression-induced recovery, and other divergence cases.

For reconstruction and summarization tasks, we evaluate semantic preservation using BLEU, ROUGE, and BERTScore. Specifically, we report ROUGE-1, ROUGE-2, and ROUGE-L, as well as BERTScore, each computed in terms of precision, recall, and F1. These metrics are applied to both reconstructed prompts versus original prompts and compressed prompts versus original prompts. To assess downstream utility preservation, we re-evaluate reconstructed prompts using LLaDA and compute exact-match accuracy against the outputs produced from the original prompts.

\section{Results}
\subsection{Autoregressive vs. Diffusion Large Language Models Under Prompt Compression}
\begin{table*}[t]
\centering
\caption{Performance comparison across reconstruction and summarization for datasets (R = recall, P = precision, F1 = F1 score).}
\label{tab:results}
\resizebox{\textwidth}{!}{
\renewcommand{\arraystretch}{1.8}
\begin{tabular}{l l c c c c c c c c c c c c c}
\hline
\textbf{Dataset} & \textbf{Task} & \textbf{BLEU} & \textbf{R1-F1} & \textbf{R1-P} & \textbf{R1-R} & \textbf{R2-F1} & \textbf{R2-P} & \textbf{R2-R} & \textbf{RL-F1} & \textbf{RL-P} & \textbf{RL-R} & \textbf{BERT-P} & \textbf{BERT-R} & \textbf{BERT-F1} \\
\hline
GSM8K & Reconstruction 
& 0.2641 & 0.5577 & 0.6534 & 0.5191 
& 0.3558 & 0.4126 & 0.3306 
& 0.5152 & 0.6064 & 0.4768 
& 0.9441 & 0.9291 & 0.9363 \\

ShareGPT & Summarization 
& 0.2396 & 0.5053 & 0.5218 & 0.5026 
& 0.2798 & 0.2899 & 0.2757 
& 0.3922 & 0.4053 & 0.3905 
& 0.9114 & 0.9087 & 0.9098 \\

DUC2004 & Summarization 
& 0.2678 & 0.5449 & 0.5971 & 0.5315 
& 0.3291 & 0.3606 & 0.3130 
& 0.5016 & 0.5507 & 0.4887 
& 0.9490 & 0.9394 & 0.9440 \\
\hline
\end{tabular}
}
\end{table*}
\begin{table}[t]
\caption{Performance comparison between autoregressive models and DLLMs.}
\label{tab:ar_vs_dllm}
\centering
\begin{tabular}{llllr}
\toprule
Task & Dataset & Metric & AR \cite{zhang2025empirical} & DLLM \\
\midrule
Reconstruction & GSM8K & BLEU & 0.55 & 0.26 \\
Reconstruction & GSM8K & ROUGE-L-F1 & 0.86 & 0.52 \\
Reconstruction & GSM8K & BERTScore-F1 & 0.94 & 0.94 \\
Summarization & DUC2004 & BLEU & 0.34 & 0.27 \\
Summarization & DUC2004 & ROUGE-L-F1 & 0.21 & 0.50 \\
Summarization & DUC2004 & BERTScore-F1 & 0.85 & 0.94 \\
Math & GSM8K & Agreement$^{*}$ & 0.22 & 0.24 \\
\bottomrule
\multicolumn{5}{p{0.9\linewidth}}{\small
Agreement$^{*}$ — The answers to the original prompt and the compressed prompt both match the ground truth.
}
\end{tabular}
\end{table}

Table~\ref{tab:ar_vs_dllm} compares compressed-prompt performance across autoregressive large language models and diffusion large language models on reconstruction, summarization, and math reasoning. Further test results for the diffusion large language model are provided in Table~\ref{tab:results}.

The results show that prompt compression behaves differently across architectures and tasks. For GSM8K reconstruction, the autoregressive model performs better on BLEU (0.55 vs 0.26) and ROUGE-L (0.86 vs 0.52), which mainly reflect surface-level overlap with the original text. However, both models achieve the same BERTScore (0.94), indicating that even though the DLLM reproduces fewer exact words, it preserves the underlying meaning just as well.

For DUC2004 summarization, the pattern is less consistent. The DLLM is better on ROUGE-L (0.50 vs 0.21) and BERTScore (0.94 vs 0.85), suggesting better preservation of key content and meaning. However, the autoregressive model has a slightly higher BLEU (0.34 vs 0.27), which may indicate it stays closer to the reference wording, even if it misses or distorts some of the important content. For GSM8K math agreement, the DLLM score is slightly higher (0.24 vs 0.22), meaning it more often produces answers that match the ground truth in both original and compressed conditions. This difference is small, so it should be interpreted as a weak trend rather than a strong claim. Instead, the table is intended to situate the observed LLaDA behavior relative to previously reported autoregressive compression outcomes. Overall, prompt compression does not affect all evaluation metrics in the same way, and different model families respond differently depending on whether the metric emphasizes wording or meaning.

\subsection{Math Outcome Partition Analysis}
To understand how compression affects GSM8K math reasoning, we partitioned cases by whether the ground-truth answer matched the original-prompt response, the compressed-prompt response, both, or neither. “Correct” and “Incorrect” indicate whether each response matched the GSM8K ground-truth numerical answer, while “Orig. vs. Comp.” indicates whether the original-prompt and compressed-prompt answers were identical to each other. These results are summarized in Table~\ref{tab:math_partitions}.

The largest group is where both the original and compressed prompt responses are incorrect and different. This group is 35.6\% of the total dataset, meaning compression most often changed the model's failure trajectory rather than simply preserving the original response. These cases also have the highest prompt complexity, including longer prompts, more tokens, and higher numerical density. The group where only the original prompt response is correct accounts for 27.6\% of cases, showing that compression sometimes removes information needed for correct reasoning. However, because these prompts are similar in length to prompts where both responses are correct, these failures are likely due to loss of key semantic relations rather than length alone. The group where both prompt responses are correct makes up 24.4\% of cases and contains the simplest prompts overall, suggesting that compressed prompts are most reliable on lower-complexity arithmetic problems. The group where only the compressed prompt response is correct is only 1.2\%, so cases where compression improves math accuracy are rare and should not be overinterpreted.

\subsection{Task-Specific Sensitivity to Prompt Compression}
Prompt compression affects LLaDA differently across reconstruction, summarization, and mathematical reasoning tasks. Although semantic similarity metrics (especially BERTScore) remain relatively high across all datasets, downstream task performance shows that some task types are significantly more sensitive to compression than others. 
Summarization task outputs are the most stable under compression. DUC2004 and ShareGPT show the highest semantic preservation, with BERTScore F1 values of 0.9440 and 0.9098 respectively. Despite compression rates of approximately 46--47\%, LLaDA generally preserves the semantic content of its original-prompt outputs when generating from compressed prompts. However, BLEU and ROUGE are noticeably lower, suggesting that while the overall meaning is preserved, surface-level phrasing and token-level structure are altered. This indicates that summarization is relatively robust to compression, likely because it depends more on global discourse meaning than exact token sequences. 


For GSM8K reconstruction, it achieves a BERTScore F1 of 0.9363, indicating that LLMLingua-2 often retained the overall meaning of the original prompts even after compression. However, lexical overlap metrics were substantially lower, with BLEU at 0.2641 and ROUGE-L F1 at 0.5152. This suggests that reconstructed prompts frequently preserved semantics while altering wording, phrasing, or structural details. Across GSM8K and DUC2004, BERTScore recall was consistently lower than precision: for GSM8K reconstruction, precision and recall were 0.9441 and 0.9291 respectively, while DUC2004 obtained 0.9490 precision and 0.9394 recall. This suggests that compression tends to remove information rather than introduce incorrect content, leading to mild information loss rather than semantic drift. 

Mathematical reasoning is far more sensitive to these omissions. Although reconstructed GSM8K prompts remained semantically similar to the originals, accuracy degrades considerably when compressed prompts are used for inference. Responses generated from original prompts aligned with the ground-truth answer substantially more often (52.0\%) than responses generated from compressed prompts (25.6\%). Furthermore, when reconstructed prompts are fed back into LLaDA and compared directly against the original-prompt responses, agreement is only 35.2\%. This aligns with the partition analysis in Section~\ref{tab:math_partitions}, where the most frequent outcome class contains cases where both the original and compressed prompts produce different incorrect answers. Together, these results indicate that preserving general semantic similarity is not sufficient for maintaining reasoning in mathematical tasks.


\begin{table}[t]
\caption{GSM8K math performance partitions: answer correctness and original/compressed consistency.}
\label{tab:math_partitions}
\centering
\begin{tabular}{lllr}
\toprule
Original & Compressed & Orig. vs. Comp. & \% \\
\midrule
Incorrect & Incorrect & Different & 35.6 \\
Correct & Incorrect & Different & 27.6 \\
Correct & Correct & Same & 24.4 \\
Incorrect & Incorrect & Same & 11.2 \\
Incorrect & Correct & Different & 1.2 \\
\bottomrule
\multicolumn{4}{@{}p{\linewidth}}{\small
$^*$ \% is percentage out of 250 samples.
}
\end{tabular}
\end{table}

\section{Discussion}
The results suggest that prompt compression in DLLMs is not merely a matter of semantic preservation. Although compressed prompts often remain semantically similar to the originals, downstream behavior changes substantially, especially in mathematical reasoning tasks. This implies that DLLMs may use prompt information differently from autoregressive models during iterative denoising. A possible explanation is that DLLMs are sensitive to how conditioning information is distributed across the prompt. Thus, even small omissions can disrupt denoising by removing information that helps stabilize multiple token predictions, causing semantically similar prompts to follow different reasoning trajectories.


The precision-recall imbalance observed in BERTScore provides insight into the nature of compression failures. Since recall is consistently lower than precision, the compressed or reconstructed prompts more often omitted information than introduced unrelated content. This distinction is important because it suggests that the primary issue is not semantic corruption, but insufficient conditioning information. In diffusion-based generation, omitted entities, quantities, or relational cues may remove anchors that help guide iterative token refinement, making the generation process less stable even when the remaining prompt still appears semantically coherent.

These findings suggest that prompt compression methods optimized for autoregressive models may not directly transfer to a diffusion architecture. LLMLingua-2 estimates token importance using bidirectional contextual signals, but its evaluation and optimization context are still dominated by autoregressive downstream models. Tokens that appear redundant under sequential decoding may still contribute globally useful conditioning information in diffusion models. This raises the possibility that diffusion-aware compression strategies may need to consider denoising stability, token interaction patterns, or global conditioning effects rather than relying only on autoregressive saliency estimates. More broadly, the study highlights limitations in evaluating prompt compression primarily through semantic similarity metrics such as BERTScore or ROUGE. High semantic preservation does not necessarily imply stable downstream reasoning behavior. For DLLMs in particular, future evaluation frameworks may need to incorporate behavioral consistency, reasoning stability, or trajectory-level analyses in addition to semantic overlap measures.

\section{Conclusion}
This study evaluates the transferability of LLMLingua-2 prompt compression to the diffusion large language model LLaDA across reconstruction, summarization, and mathematical reasoning tasks. The results show that compression can preserve broad semantic meaning, especially for summarization and reconstruction, but this does not guarantee stable downstream behavior. Mathematical reasoning is particularly sensitive to compression, suggesting that omissions in prompt information can significantly affect reasoning outcomes. These findings suggest that prompt compression methods developed for autoregressive models do not transfer uniformly to diffusion language models. Future work should develop diffusion-aware compression methods that preserve not only semantic similarity but also task-critical information needed for stable denoising and reasoning.

\section{Limitations and Future Work}
This work evaluates one diffusion large language model, LLaDA, and one prompt compression framework, LLMLingua-2. The experiments are also limited to three datasets covering reconstruction, summarization, and mathematical reasoning. Additional diffusion architectures, compression methods, datasets, and reasoning domains may reveal different transfer behaviors and provide a more complete understanding of prompt compression in DLLMs.

Future work could develop prompt compression techniques specifically designed for DLLMs. Such methods could better account for the bidirectional attention mechanism and iterative denoising process used during diffusion-based generation. In particular, DLLM-specific compressors may aim to preserve tokens that provide globally important conditioning signals across denoising steps, even when those tokens appear less critical under conventional autoregressive compression criteria. Future studies could also investigate token importance estimation strategies that reflect denoising stability, reasoning consistency, and the preservation of relational or numerical information. Rather than optimizing compression only for semantic similarity or reduced prompt length, diffusion-aware compression methods may explicitly model how removed information affects iterative refinement during generation. This could lead to adaptive compression strategies that retain task-critical prompt structures for reasoning-intensive tasks while applying stronger compression to more redundant content.

\begin{acks}
This work was supported by the Mitacs Globalink Research Award and the Innovative Cognitive Computing Research Center.
\end{acks}

\section*{AI Usage Declaration}
An AI assistant was used for grammar checking and paraphrasing to improve the clarity and readability of the manuscript.
\bibliographystyle{ACM-Reference-Format}
\bibliography{sample-base}

\end{document}